\documentclass[letterpaper, 10 pt, conference]{ieeeconf}  

\IEEEoverridecommandlockouts                              

\usepackage{times}
\usepackage{amsfonts} 
\usepackage{graphicx}
\usepackage{tabularx}
\usepackage{comment}
\usepackage{wrapfig}
\usepackage{amsmath}
\usepackage{amssymb}
\usepackage{array}
\usepackage{wrapfig,lipsum,booktabs}
\usepackage{hyperref}
\usepackage{booktabs}
\usepackage{multirow}
\usepackage{subfig}
\usepackage{setspace}
\usepackage{xcolor}

\usepackage{algorithm}
\usepackage{algorithmic}

\usepackage{hyperref}

\makeatletter
    \let\NAT@parse\undefined
\makeatother
\usepackage[square,numbers,sort&compress]{natbib}
\setcitestyle{numbers}
\usepackage{multicol}

\usepackage{titlesec}

\usepackage{titlesec}

\begin{document}


\title{\LARGE \bf
R+X: Retrieval and Execution from Everyday Human Videos

}

\author{Georgios Papagiannis*,  Norman Di Palo*,  Pietro Vitiello and Edward Johns \footnote{lets' gooo}\vspace{-1ex}}

\twocolumn[{
\renewcommand\twocolumn[1][]{#1}
\maketitle
\begin{center}
    \centering
    \captionsetup{type=figure}

  \includegraphics[width=.9\linewidth]{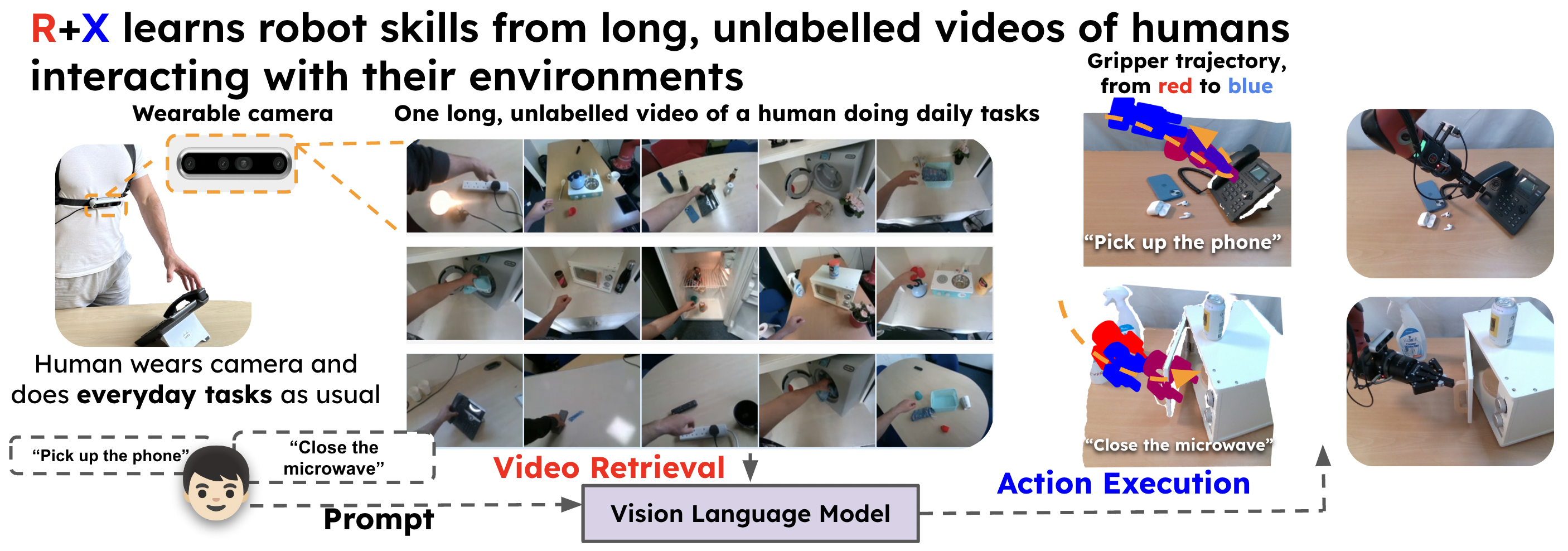}
    \caption{Given a language prompt, R+X first retrieves short relevant video clips extracted from a long unlabelled video of a human performing everyday tasks, recorded with a wearable camera. By using the retrieved video clips and a VLM, R+X performs in-context imitation learning, allowing it to immediately generate and execute the desired behaviour on the robot.}
    \label{fig:figure1}
\end{center}
}]

\thispagestyle{empty}
\pagestyle{empty}

\begin{abstract}
We present R+X, a framework which enables robots to learn skills from long, unlabelled, first-person videos of humans performing everyday tasks. Given a language command from a human, R+X first retrieves short video clips containing relevant behaviour, and then executes the skill by conditioning an in-context imitation learning method (KAT) on this behaviour. By leveraging a Vision Language Model (VLM) for retrieval, R+X does not require any manual annotation of the videos, and by leveraging in-context learning for execution, robots can perform commanded skills immediately, without requiring a period of training on the retrieved videos. Experiments studying a range of everyday household tasks show that R+X succeeds at translating unlabelled human videos into robust robot skills, and that R+X outperforms several recent alternative methods. Videos and code are available at \href{https://www.robot-learning.uk/r-plus-x}{https://www.robot-learning.uk/r-plus-x}.
\end{abstract}

\let\thefootnote\relax\footnotetext{$^*$ Joint first authorship. The Robot Learning Lab at Imperial College London. Contact at: \texttt{ \{g.papagiannis21, n.di-palo20\}@imperial.ac.uk}}


\section{Introduction}


In this work, we study the problem of learning robot skills from long, unlabelled, first-person videos of humans performing everyday tasks in everyday environments. \textit{"Long, unlabelled videos"} means that a human simply goes by their everyday life and passively records videos of themselves without the need to specify which behaviour is being performed. \textit{"Everyday environments"} means that videos contain diversity in tasks, scenarios, objects, distractors, illumination, and camera viewpoints and motion. It is likely that such videos will become abundant through the adoption of wearable devices such as AR headsets and glasses \cite{visionpro, metaquest, rayban, magicleap}, and thus offer an unprecedented opportunity for future scalability if robots could learn from such videos, instead of learning exclusively from costly tele-operation.

Previous approaches to learning from videos of humans have often relied on a set of strong constraints and assumptions, such as human videos manually aligned with robot videos, human videos manually labelled with language descriptions or demonstrations on robot or MoCap hardware \cite{jain2024vid2robot, whirl, jang2022bcz, heppert2024ditto, bahety2024screwmimic, zhu2024orion, wang2024dexcap}. In this work, we remove all of these constraints completely and present a framework that requires only an unlabelled first-person video depicting tens of tasks.

Our framework, \textbf{R+X}, is a two-stage pipeline of \textbf{R}etrieval and E\textbf{X}ecution shown in Figure~\ref{fig:figure1}, which uses Foundation Models for both stages. Upon receiving a language command from a user, a Vision Language Model retrieves all clips where the human executes the specified task. By extracting the trajectories of the human’s hand in each clip, we then employ a few-shot in-context imitation learning method to condition on these trajectories, which enables a robot to ingest, learn from, and replicate the retrieved behaviours to previously unseen settings and objects. The recent literature demonstrated that, by finetuning large Vision Language Models on robotics data, they can transfer their common knowledge and ability to reason and plan to robotics settings \cite{brohan2023rt2, driess2023palme}. This, however, requires very expensive finetuning of often intractably large models. With our proposed framework, we can equally leverage these abilities but now via retrieval and video understanding, thus without the need for any finetuning. 

In summary, the two main properties of R+X are: \textbf{1)} it enables robots to execute everyday tasks from language commands, given long, unlabelled, first-person videos collected naturally by recording a human's daily activities, and \textbf{2)} it achieves this without the need for any training or finetuning of models, allowing it to learn and execute tasks immediately, while still inhereting many strengths of large Vision Language Models.

\vspace{-1ex}\section{Related Work}
\label{sec:related}

\begin{figure}[t]
    \begin{center}
\vspace{1ex}   \includegraphics[width=0.4\textwidth]{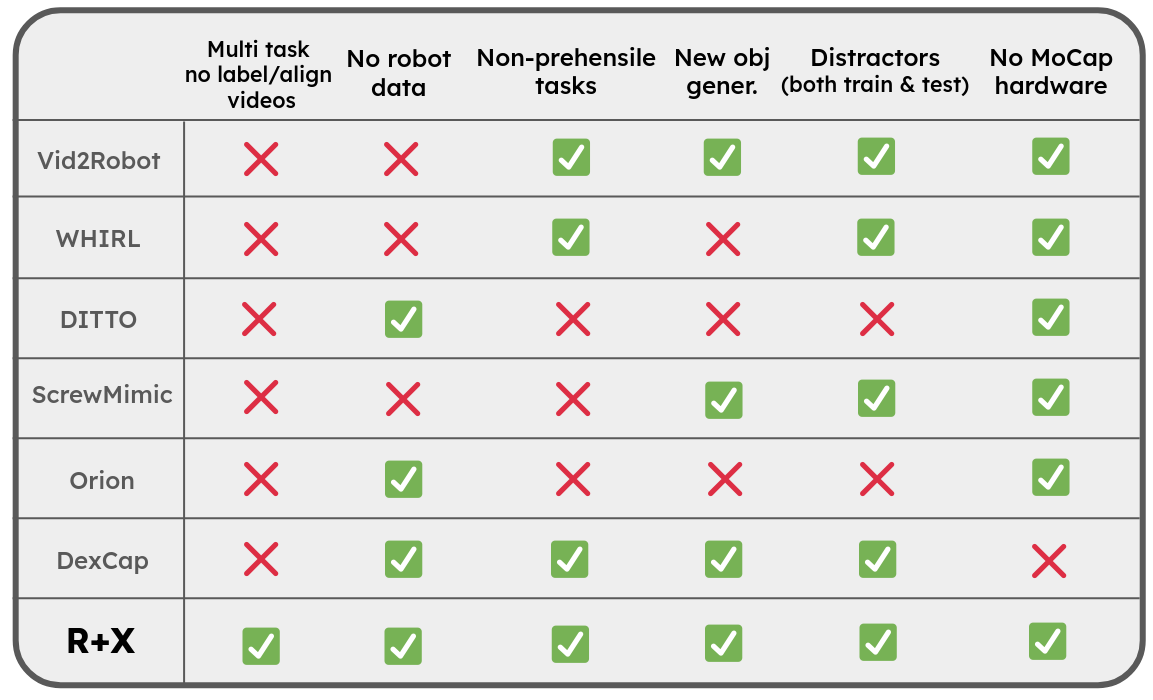}
    \caption{The main assumptions and constraints of many recent Learning from Observation methods.}
    \label{fig:taxonomy}

    \end{center}

      \vspace{-6ex}

\end{figure}

\textbf{Collecting Robotics Data.} Scaling data collection has been proven to be a successful path towards increasingly more general machine learning models \cite{brown2020language, geminiteam2023gemini, general-pattern-machines, chinchilla}. To collect robotics data, the most common paradigm is to teleoperate robots, collecting datasets of paired observations and actions \cite{embodimentcollaboration2023open}. This, however, needs dedicated teleoperation hardware, and needs human operators to allocate their time to \textit{actively teach a robot new tasks}. In our framework, a human user interacts with their environments as usual, completing the tasks they wish, while a robot \textit{passively learns to emulate it}, resulting in a more scalable and time efficient paradigm.

\textbf{Learning from Human Videos.} Many recent works have proposed solutions to teach new skills to robots by observing a human executing such tasks. However, as we illustrate in Fig. \ref{fig:taxonomy}, they often relied on a set of assumptions and constraints that limited their use "in-the-wild". \cite{whirl, wang2023mimicplay} require a combination of human data and either robot exploration or teleoperation, therefore needing active assistance of the user in teaching the robot, and videos are recorded from a fixed, third-person camera. R+X relies entirely on the videos recorded by the user with a mobile camera in their natural environments. \cite{heppert2024ditto, zhu2024orion, bahety2024screwmimic} can learn robot skills entirely from human videos. \cite{heppert2024ditto, zhu2024orion}, however, focus on replicating the object trajectory from the demo, and cannot perform tasks that do not involve grasping and moving, such as pushing or pressing, that R+X can execute. \cite{bahety2024screwmimic} can learn a larger repertoire of skills, but still relies on learning single tasks in isolation from a fixed camera. Instead, R+X does not assume a fixed camera and can replicate tasks from a given language command after receiving a single, long, unlabelled video depicting tens of tasks, without the need to specify which clip demonstrated which behaviour. Methods like \cite{wang2024dexcap} allow a user to naturally interact with their environment while collecting data that can teach a robot new skills. However, they require additional hardware to wear, like specialised MoCap gloves. R+X only needs a single RGB-D camera, and does not require any extrinsics calibration. This means that data could be recorded using a wearable camera, smart glasses, AR visors, and more. 

\textbf{Language-Conditioned Policy Learning.} Language is considered a viable way to instruct robots and guide their task execution \cite{embodimentcollaboration2023open, driess2023palme, brohan2023rt2}. The common approach is to train a large, language and image conditioned policy that can, at test time, receive user commands and camera observations \cite{embodimentcollaboration2023open}. This, however, presents some criticalities: training such models can require enormous computational effort \cite{brohan2023rt2}, and unlike other machine learning applications, robots should learn new skills over time. Therefore, re-training or finetuning such networks should be avoided. R+X therefore does not train or finetune any networks, but takes advantage of pre-trained models able to retrieve examples of tasks \cite{geminiteam2023gemini}, and to learn to emulate and execute new behaviour directly at deployment via in-context learning \cite{KAT}. 

{Additional papers are discussed in the appendix.}


\begin{figure*}[t!]
    \centering
    \includegraphics[width=.9\textwidth]{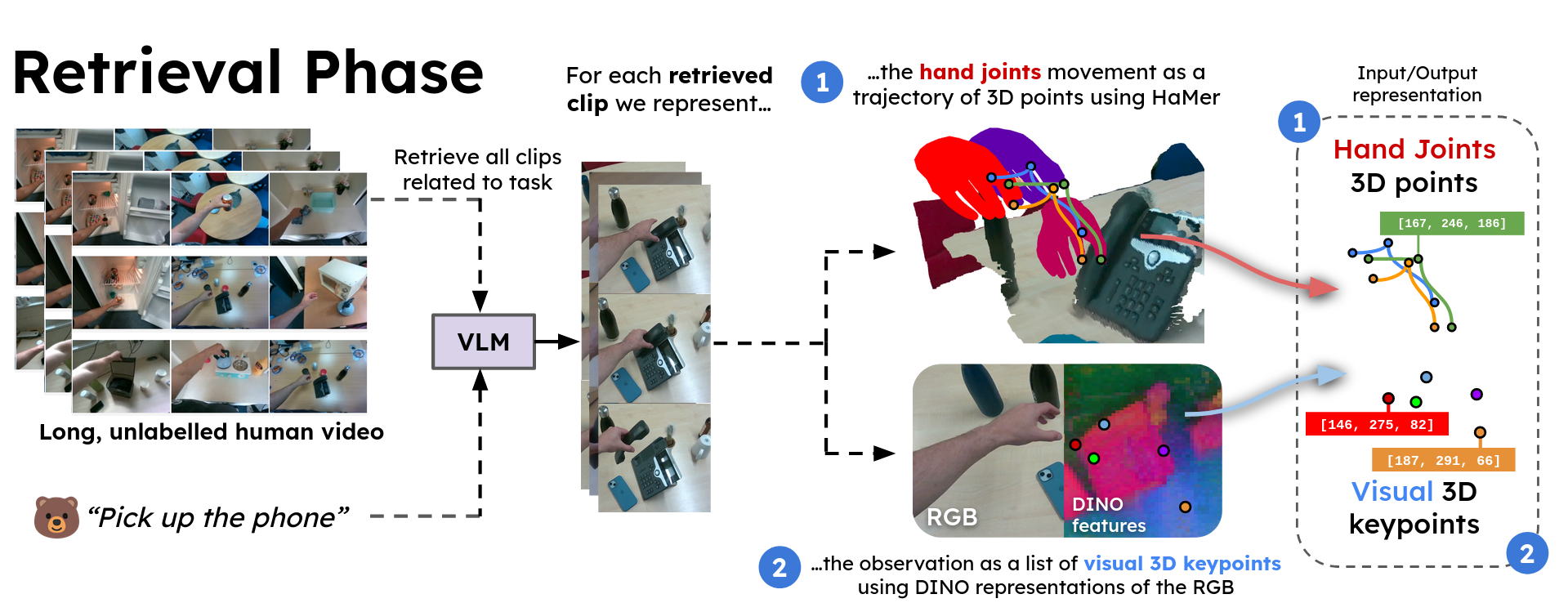}
    \centering
\vspace{-1ex}    \caption{Upon receiving a language command, R+X retrieves all the relevant clips from the human video. Each retrieved clip is transformed from pixels to a sparse 3D points representations of the hand joints movement and salient parts of the visual observation.}
    \label{fig:retrieval}
          \vspace{-10pt}

\end{figure*}

\section{R+X: Retrieval and Execution}
\label{sec:method}

We now describe R+X, our proposed method to learn robot skills from long, unlabelled videos of humans interacting with their environments. Additional details accompanying our method can be found in the appendix and on our website. We assume R+X has access to a long, unlabelled video of a person performing a multitude of tasks in many different locations. We call this long, first-person video of everyday tasks the "\textit{human video}" $\mathcal{H}$ in the rest of the paper. The goal of our method is to learn to emulate the behaviours recorded in the human video upon receiving a language command. 

Overall, from a high-level perspective, our pipeline takes three inputs: \textbf{1)} a single, long, unlabelled video of all the interactions recorded by the humans, $\mathcal{H}$ \textbf{2)} a task to execute in language form, $\mathcal{L}$ and \textbf{3)} the current observation of the robot as an RGB-D image, $\mathcal{O}_{live}$. It then outputs a trajectory of 6-DoF gripper poses, which are executed by the robot to tackle the task at hand. 

There are however a set of non-trivial challenges: while receiving a language command to execute at deployment, the robot receives \textbf{no language information before deployment}: the recorded video contains no more information than the recorded visual frames. Additionally, the robot receives \textbf{no action information}: unlike the case of teleoperation, no joints or end-effector position/velocity/acceleration data are recorded, and the correct movements need to be inferred by the videos alone: the problem is additionally complex due to the cross-embodiment between human videos and final robot actions. Furthermore, as the user is interacting with their natural environment, the visual observations can be \textbf{filled with distractor objects} typical of household and offices, unrelated to the task the user is performing. To tackle these challenges, we leverage the abilities of Foundation Models.

Specifically, at the first phase of R+X, which we call the \textbf{retrieval phase}, we use a VLM to extract from the human video all the examples of the desired behaviour described in the prompt $\mathcal{L}$ as a list of $Z$ shorter video clips $[\mathcal{V}_1, \mathcal{V}_2, \dots, \mathcal{V}_Z]$. We map the $Z$ videos into a lower dimensional 3D representation, extracting for each video: (1) a list of $K$ \textit{visual 3D keypoints} that describe the scene, $[k_1, k_2, \dots, k_K]$ where $k = (x_k, y_k, z_k)$, and (2) the movement of the user's hand as a trajectory of length $T$ of 21 hand joints that parametrise the MANO hand model \cite{MANO}, $\mathcal{J} = [j_1, j_2, \dots, j_T]$, where $j = [[x_0,y_0,z_0], \dots, [x_{21}, y_{21}, z_{21}]]$. Finally, at the second stage of R+X, which we call the \textbf{execution phase}, to emulate the behaviours observed in the retrieved video clips, we condition a few-shot in-context imitation learning model on this data that, given a live observation of the environment, it generates a trajectory of 3D hand joints to execute the desired behaviour described in the prompt. To map such joints to gripper poses, we develop a custom procedure that we describe in the appendix.

\begin{figure*}[t!]
    \centering
\vspace{0.5ex}    \includegraphics[width=.9\textwidth]{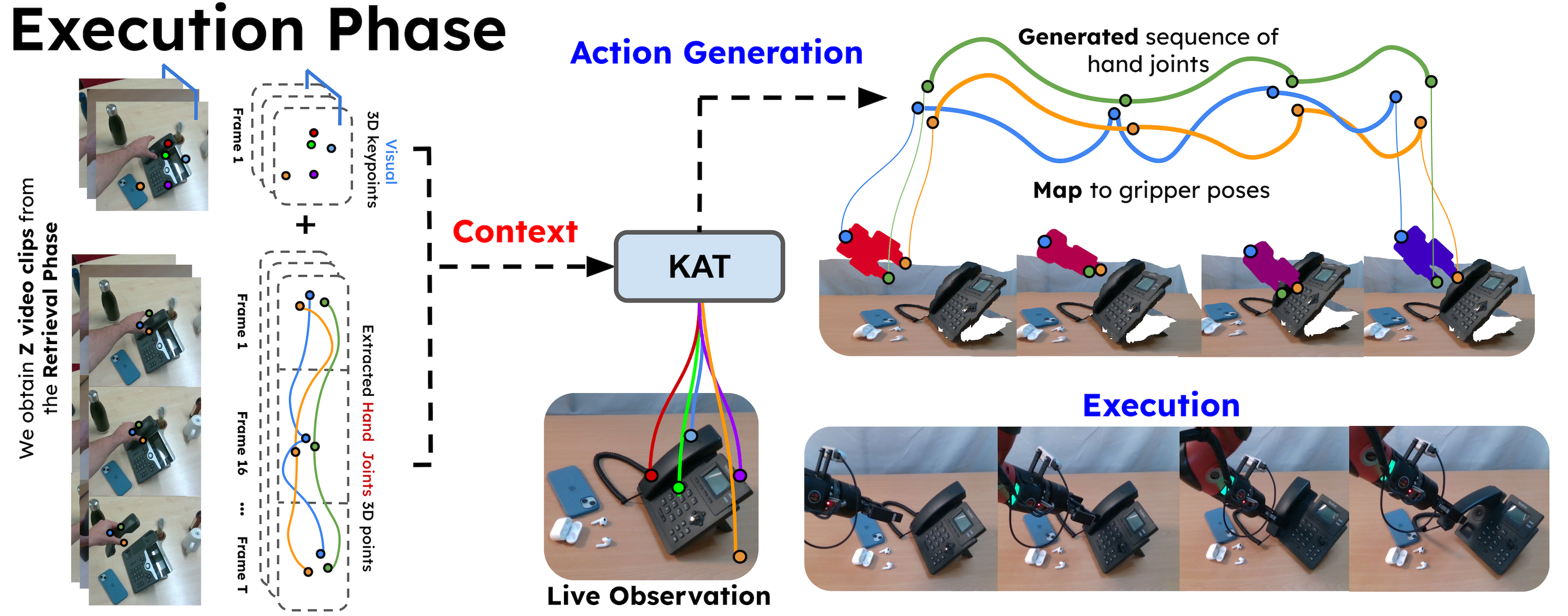}
    \centering
    \caption{The visual 3D keypoints of the first frame of each of the Z videos obtained from the retrieval phase along with each extracted hand joint trajectory are used as context for KAT. To execute a skill, visual 3D keypoints are extracted from the live observation and used as input to KAT which generates a sequence of hand joints. By mapping this sequence to gripper poses the robot executes the desired task.}
    \label{fig:execution}
    \vspace{-10pt}
\end{figure*}

\subsection{Retrieval: Extracting visual examples from a long, unlabelled video}\label{sec:retrieval} 
The first main phase of R+X is \textbf{retrieval}, shown in Figure~\ref{fig:retrieval}. Upon receiving a language command $\mathcal{L}$, and given the human video $\mathcal{H}$, the goal is to retrieve all video clips from the human video that depict the execution of the requested task. This is accomplished using a recent Vision Language Model (VLM), Gemini Pro 1.5 Flash \cite{gemini15}. Gemini, which we denote $\mathcal{G}$, is natively multi-modal and can take as input images, videos, and text, and outputs text. We prompt the model with the human video, and ask it to retrieve the starting and ending seconds at which the received task happens. The inputs of this phase is therefore a language command and the human video, and the output is a list of $Z$ shorter video clips demonstrating the desired task, $\mathcal{G}(\mathcal{H}, \mathcal{L}) \rightarrow [\mathcal{V}_1, \dots, \mathcal{V}_Z]$. Each clip comprises $T$ RGB-D frames, where $T$ can vary across different clips. 


\subsubsection{Preprocessing Videos into a Sparse 3D Representation}

 Given the $Z$ extracted video clips, $[\mathcal{V}_1, \dots, \mathcal{V}_Z]$, we apply a preprocessing step that converts each video clip from a list of RGB-D frames to a set of 3D points describing the visual scene and the trajectory of the human's hand, a representation that we will then feed to our few-shot in-context imitation learning model.

\textbf{\textit{Visual 3D keypoints}.} To transform the complex RGB-D observations into a lower-dimensional, easier to interpret input, we harness the powerful vision representations generated by DINO \cite{caron2021dino}, a recent Vision Foundation Model. As proposed in \cite{amir2022dinokp, KAT}, given the $Z$ clips retrieved as described before, we find a set of $K$ common 3D visual keypoints that capture interesting semantic or geometrical aspects of the scene. We extract these keypoints from the \textit{first frame} of each of the $Z$ videos only.

We first compute the DINO descriptors for the first frame of each of the $Z$ videos, obtaining a list of $Z$ different $N \times 384$ outputs, where $N$ is the number of patches of each image \cite{dosovitskiy2021vit, openai2023gpt4, caron2021dino}.  
Via a clustering and matching algorithm \cite{amir2022dinokp}, we select from the $N \times 384$ descriptors of the first frame of the first video, $\mathcal{O}_{\mathcal{V}_1,1}$, a list of the $K$ descriptors that are the most common in all the remaining $Z-1$ frames. We denote these descriptors as $\mathcal{D} \in \mathbb{R}^{K \times 384}$. Therefore, this way, we autonomously extract descriptors that focus on the object of interest, as its appearance is common among videos, while distractors and overall scene will vary \cite{amir2022dinokp, KAT}.

Finally, for each of the $K$ descriptors in $\mathcal{D}$, we extract keypoints by finding the $K$ nearest neighbours between the $N \times 384$ descriptors of each the $Z$ frames, and compute their 2D coordinates. We then project these in 3D using each frame's depth image and known camera intrinsics. As such, given the retrieved clips, we obtain and store a list $\Lambda=[\mathcal{K}_{V_1}, \dots, \mathcal{K}_{V_z}]\in\mathbb{R}^{Z\times K\times 3}$ of visual 3D keypoints, $K$ for each clip, where each $\mathcal{K}_{V_i}=[k_1,..., k_K]$ and $k_j=(x_j, y_j, z_j)$. For a more detailed description please refer to \cite{amir2022dinokp}.






\textbf{\textit{Hand Joint Actions}.} To extract human actions from each retrieved video clip we use the HaMeR model \cite{HaMeR}, a recent technique for 3D hand pose estimation from images based on DINO. In particular, using HaMeR we extract from each video frame $\mathcal{O}_{\mathcal{V}_z,t}$ (where $1\leq t \leq T$) of each of the $Z$ clips the 3D hand pose, represented as a set of 21 3D points, $j_{\mathcal{V}_z,t}$, describing the hand joints that parameterise the MANO hand model, as it is commonly done in the literature \cite{MANO}. As the camera moves between frames, we design a stabilisation technique to compute transformations between camera poses that is robust to dynamic scenes (for more details please see the appendix). This enables us to express the extracted hand joints relative to a single reference frame, that of the first camera frame of each clip. For each video clip $\mathcal{V}_z$, this process results in a sequence of 3D hand joint actions  $\mathcal{J}_{\mathcal{V}_z} = [j_{\mathcal{V}_z,1}, \dots, j_{\mathcal{V}_z,T}]$, expressed relative to the first frame of each video $\mathcal{O}_{\mathcal{V}_z, 1}$ where the visual 3D keypoints are also expressed in. As such, at the end of this process we are left with a list $\mathcal{M}=[\mathcal{J}_{\mathcal{V}_1},... \mathcal{J}_{\mathcal{V}_Z}]$ of $Z$ hand joint action sequences. 

In summary, from the $Z$ retrieved video clips, we extracted the list of hand joints actions $\mathcal{M}$, along with the list of visual 3D keypoints $\Lambda$, that will be used as context for our in-context imitation learning model, as $Z$ input-output pairs.

\subsection{Execution: Few-Shot, In-Context Imitation from Video Examples}

The second main phase of R+X is \textbf{execution}. Our framework is based on the use of a model capable of performing few-shot, in-context imitation learning, receiving a few examples of desired inputs and outputs pairs describing a desired behaviour, and able to replicate such behaviour immediately upon receiving a new input. We use \textbf{Keypoint Action Tokens (KAT)} \cite{KAT} to achieve this, a recently proposed technique that takes 3D visual keypoints as input and outputs a trajectory of 3D points describing the gripper movement. Instead of explicitly training a model on robot data, KAT demonstrates that recent, off-the-shelf Large Language Models are able to extract such numerical patterns, and behave as few-shot, in-context imitation learning machines, without the need for any further finetuning.  

\begin{figure*}[t!]
    \centering
    \includegraphics[width=.99\textwidth]{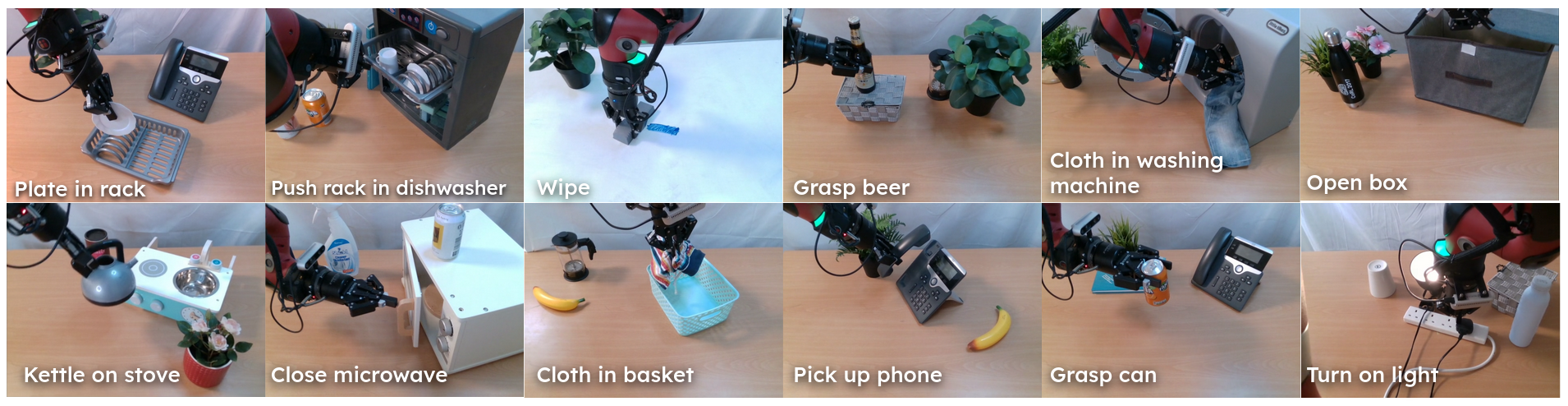}
    \centering
    \caption{We test R+X on 12 everyday tasks, executed by a human in different rooms and with different distractors.}\vspace{-3ex}
    \label{fig:tasks}
\end{figure*}

Given the output of the retrieval phase $[\Lambda, \mathcal{M}]$ and a new, live RGB-D observation collected by the robot $\mathcal{O}_{live}$, we extract its visual 3D keypoints representation $\mathcal{K}_{live}$ by first extracting its $N \times 384$ DINO descriptors, and then finding the $K$ nearest neighbours to each of the $K$ descriptors in $\mathcal{D}$, that we obtained in the retrieval phase. This results into $K$ 2D coordinates, that we project in 3D. We then input to KAT as context $[\Lambda, \mathcal{M}]$, as $Z$ examples of the desired input-output mapping, and the new visual 3D keypoints, and generate a new trajectory of desired hand joint actions, $\text{KAT}([\Lambda, \mathcal{M}], \mathcal{K}_{live}) \rightarrow \mathcal{J}_{live}$ with $\mathcal{J}_{live} = [j_{1}, \dots, j_{T}]$, as shown in Figure~\ref{fig:execution}. We then map the predicted trajectories of hand joints actions into gripper poses as described in our website and appendix, and execute them.

To summarize, given a language command $\mathcal{L}$ and a human video $\mathcal{H}$, R+X first \textbf{retrieves} $Z$ videos depicting the described behaviour using a VLM. From the retrieved videos, a list of $[\Lambda, \mathcal{M}]$ visual 3D points and hand joint actions are extracted. Then, to execute the desired behaviour described in the prompt $\mathcal{L}$, $[\Lambda, \mathcal{M}]$ along with the visual 3D keypoints of the live observation $\mathcal{K}_{live}$ are used as context for KAT that performs few-shot in-context imitation learning to generate a trajectory of hand joints actions, which is mapped to gripper poses and \textbf{executed} on the robot.


\section{Experiments}
\label{sec:result}


\textbf{Human Video.} We collect the human video $\mathcal{H}$ using an Intel RealSense 455, worn by a human on their chest as shown in Figure~\ref{fig:figure1}. To reduce downstream computational time, we filter out each frame in which human hands are not visible right after recording. As our robot is single-armed, we limit ourselves to single hand tasks. However, our method could identically be applied to bimanual settings and dexterous manipulators. \textcolor{black}{The video is collected in many different rooms and buildings.} 


\textbf{Robot Setup.} At execution, we use a Sawyer robot equipped with a RealSense 415 head-camera. The robot is equipped with a two-fingered parallel gripper, the Robotiq 2F-85. As the robot is not mobile, we setup different scenes in front of it with variations of the tasks recorded by the human, placing several different distractors for each task, while the human video was recorded in many different rooms. Although we have a wrist-camera mounted, we do not use that in our work.

\textbf{Tasks.} To evaluate our proposed framework, we use a set of 12 everyday tasks, where a human interacts with a series of common household objects, listed in Fig. \ref{fig:tasks}. We include movements like grasping, opening, inserting, pushing, pressing, and wiping. 

\textbf{Baselines.} We compare R+X, and its retrieval and execution design, to training a single, language-conditioned policy. To obtain language captions from the human video, we use Gemini to autonomously caption snippets of the video, obtaining a \textit{(observation, actions, language)} dataset.  We finetune R3M (ResNet-50 version \cite{he2015deep}) \cite{nair2022r3m} and Octo \cite{octomodelteam2024octo} on this data. We extend R3M to also encode language via SentenceBERT and use a Diffusion Policy \cite{chi2023diffusion} head to predict actions from intermediate representations. We denote this version as R3M-DiffLang. 

Videos, code, and additional implementation details can be found on our website: \href{https://www.robot-learning.uk/r-plus-x}{https://www.robot-learning.uk/r-plus-x}

\begin{table*}
\centering
\footnotesize\setlength{\tabcolsep}{8.pt}
\vspace{1ex}\begin{tabular}{l@{\hspace{6pt}} *{13}{c}}
\toprule
\cmidrule(l){1-14}
\bfseries Method / Task&  Plate&  Push&  Wipe&  Beer&  Wash &  Box&  Kettle&  Micro.& Basket & Phone & Can & Light & \textbf{Avg.}\\ 
\midrule
\bfseries R3M-DiffLang
& 0.5 & 0.7 & 0.4 & 0.7 & 0.5 & 0.5 & 0.4 & 0.8 & 0.7 & 0.4 & 0.7 & 0.3 & 0.55\\  
\midrule

\bfseries Octo
& 0.5 & 0.8 & 0.5 & 0.6 & 0.5 & 0.5 & 0.4 & 0.7 & 0.6 & 0.4 & 0.6 & 0.3 & 0.53\\ 

\midrule
\bfseries R+X
& \textbf{0.6}  & 0.8 & \textbf{0.7} & \textbf{0.8} & \textbf{0.6} & \textbf{0.7} & \textbf{0.6} & 0.8 & 0.7 & \textbf{0.7} & \textbf{0.8} & \textbf{0.6} & \textbf{0.7}\\ 

\bottomrule
\addlinespace

\end{tabular}
\vspace{-1ex}\caption{Result of the various methods on the 12 proposed tasks.}\vspace{-2ex}\label{tab:main_table}

\end{table*}

\vspace{-.5ex}\subsection{Results}
\textbf{Can R+X learn robot skills from long, unlabelled videos? How does it perform with respect to a monolithic language-conditioned policy?} In these experiments, we evaluate the performance of R+X in learning a set of everyday skills. At deployment, we place the object to be interacted with in front of the robot, and issue a language command. We then run the retrieval and execution pipeline using the human video, the current observation and the language command. We run 10 evaluations per task, randomising 1) the object pose 2) the type and number of distractors 3) for tasks where it is possible, we swap the object to interact with, with another one from the same class to test for generalisation (e.g. a different can, a different piece of clothing, a different telephone). We provide more details in the appendix.

In Table \ref{tab:main_table}, we report the performance of R+X on these tasks, together with the baselines. As we demonstrate, the framework is able to tackle a wide range of everyday tasks, surpassing the monolithic policy baselines. These results prove the benefit of modeling language-conditioned learning from observation as distinct retrieval and execution phases, to fully leverage the abilities of recent Foundation Models. \textcolor{black}{In the appendix, we investigate more in depth the performance of R+X on unseen objects and in the presence of distractors.}

\begin{figure*}[h!]
    \begin{center}
    \includegraphics[width=0.99\textwidth]{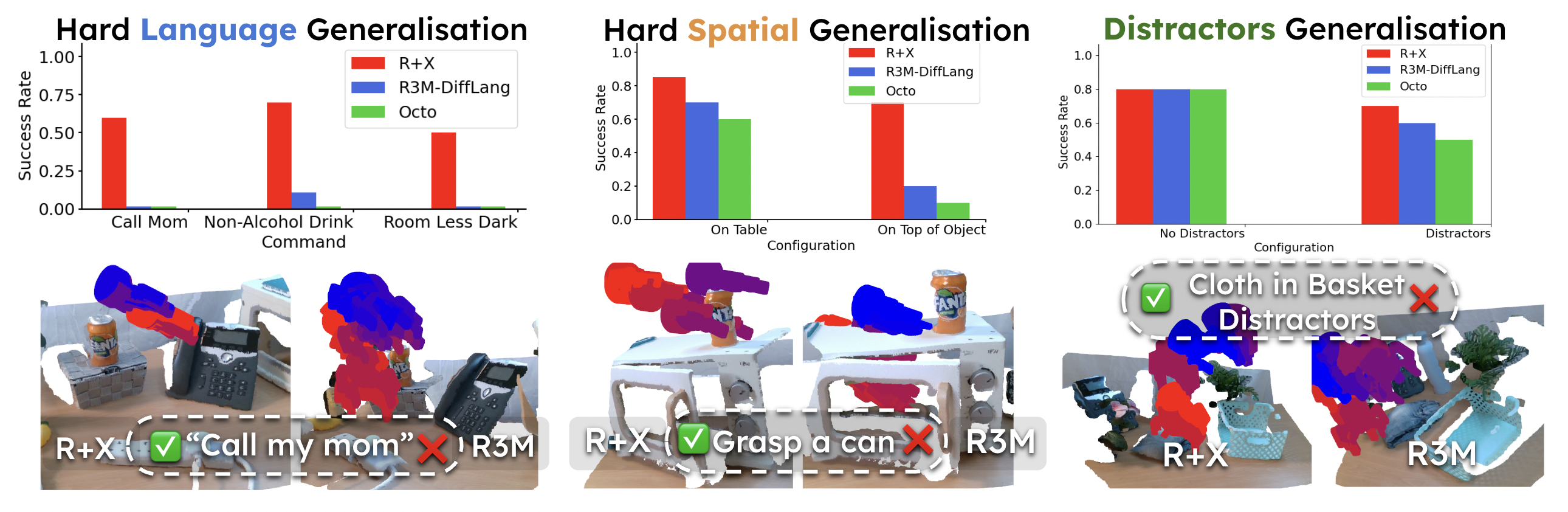}
    \end{center}\vspace{-2ex}
\caption{Examples of spatial, language and distractors generalisation. Gripper trajectories move from red to blue.}\vspace{-2.5ex}
\label{fig:hard}
\end{figure*}

\textbf{What are the main sources of difference in performance between R+X and a monolithic policy?}
In these experiments, we investigate what changes in the inputs lead to the most noticeable difference in performance between R+X and the baselines, R3M-DiffLang and Octo. We explored two aspects, related to two properties of R+X:

\textbf{\textit{Hard Spatial Generalisation}}: R+X, by retrieving videos of the desired task, can also extract a series of relevant keypoints $\mathcal{K}_{live}$ from the current observation $\mathcal{O}_{live}$, something not possible when using a single policy network. Going from RGB-D to a list of 3D points for inputs and outputs allows us to apply simple geometric data augmentation techniques for KAT, such as normalisation or random translations and rotations. This leads to stronger spatial generalisation: in the \textit{"grasp a can"} and \textit{"grasp a beer"} tasks we test each method's performance when the objects are on the table, or when they are positioned on top of other objects (a box, a microwave). By running 5 test rollouts for each case, we demonstrate how R+X retains strong performance, while the baselines' performance drops. Results and an example of the predicted gripper trajectories can be seen in Fig. \ref{fig:hard}, middle.

\textbf{\textit{Hard Language Generalisation}}: By leveraging the language and video understanding abilities of recent large Vision Language Models, such as Gemini, R+X can interpret and execute nuanced commands. To evaluate this, we setup a scene with many objects, and ask the robot to perform three tasks: \textit{"give me something to call my mom"}, \textit{"give me a non-alcoholic drink"} and \textit{"make the room less dark"}. We run 5 test rollouts for each of these commands, modifying the position of the objects and the distractors. The language-conditioned policies struggle to interpret these commands, that are strongly out of distribution. R+X, on the other hand, leverages Gemini's ability to understand the meaning of these commands, as it is able to retrieve useful video clips from the human video (respectively, picking up the phone, grasping a Fanta can, and turning on the light). Results and an example of output gripper trajectories can be seen in Fig. \ref{fig:hard}, left.

\begin{figure}[t!]
    \centering
    \includegraphics[width=.15\textwidth]{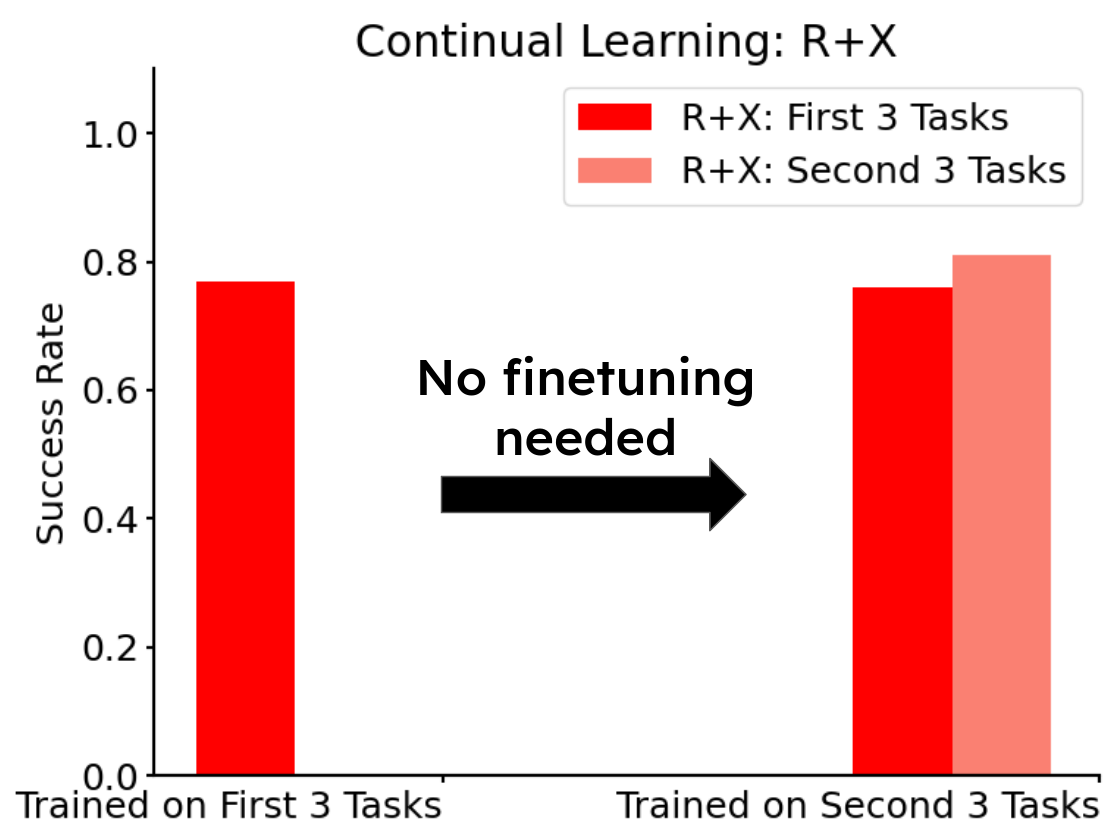}
    \includegraphics[width=.15\textwidth]{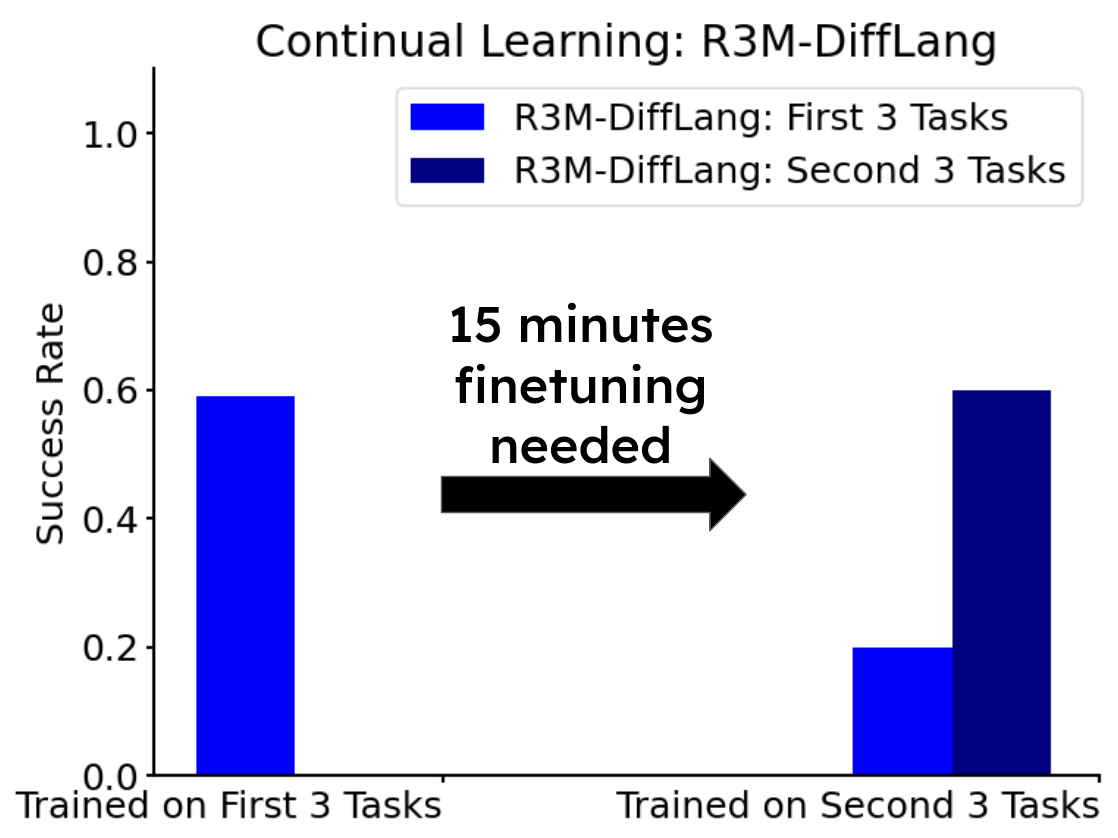}
    \includegraphics[width=.15\textwidth]{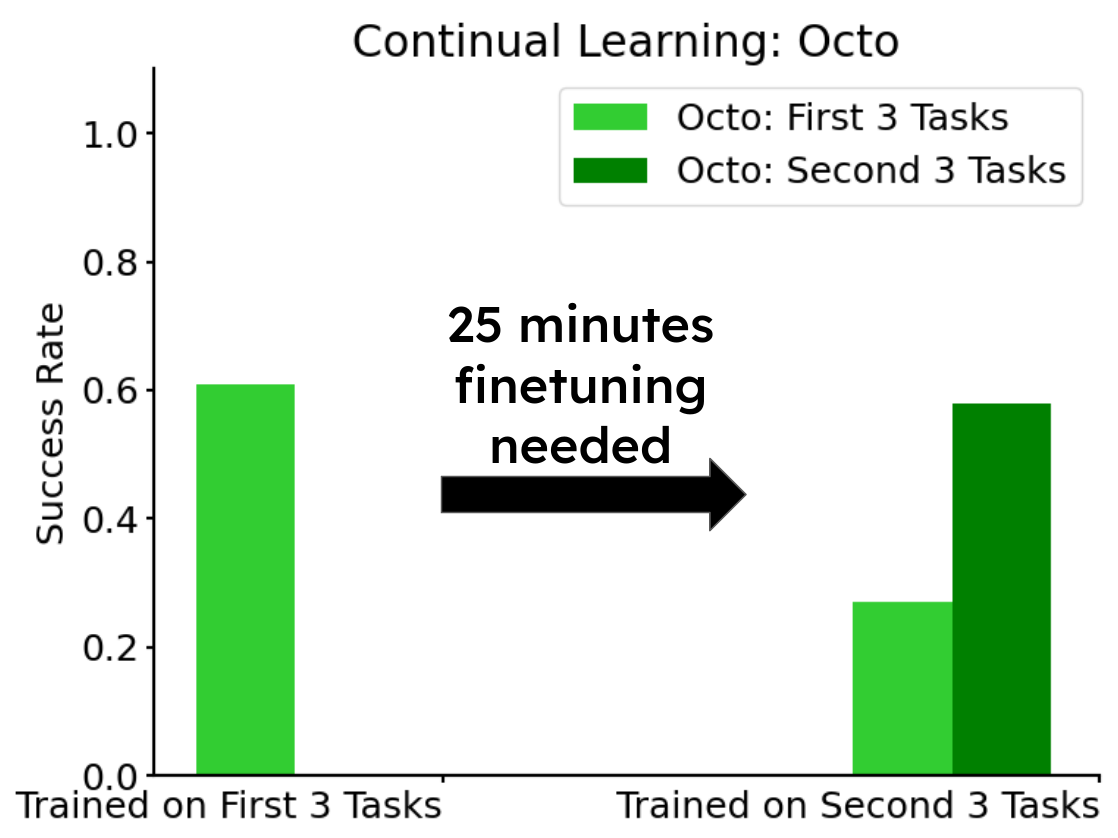}

    \centering
    \caption{We compare R+X and the baselines' ability to learn tasks in succession, and the time needed to learn such new tasks.}
    \label{fig:continual_retrieval}
\vspace{-4.5ex}
\end{figure}

\textbf{\textit{Distractors Generalisation}}: One of the main differences between R+X and the baselines is that the former, after retrieving a set of video clips from $\mathcal{H}$ depicting the requested task, can extract a set of keypoints that are semantically and geometrically meaningful for the objects to interact with, as described in the main paper. These keypoints are generally robust to distractors, as they focus only on the DINO features that are common in all images, whereas distractors typically vary throughout the human video $\mathcal{H}$ and the execution. The baselines, being monolithic policies, receive as input the observation as it is, and must learn to generalise to the presence of distractors during the long training phase.

To evaluate the effect of distractors on the scene, we test R+X and the baselines on two tasks, "\textit{grasp a can}" and "\textit{put the cloth in the basket}", emulating the experimental scenario of the "\textit{Hard Generalisations}" sections. We run 10 runs without distractors, and 10 with, explicitly measuring the performance in the two cases. We can see in the results of Fig. \ref{fig:hard} right, how R+X is more robust to the presence of random distractor objects, highlighting the advantage of extracting semantically and geometrically meaningful keypoints \textit{after} having received a language command and having retrieved the corresponding video clips.

\textbf{Can R+X learn task sequentially over time?} 
In these experiments, we demonstrate how R+X can learn tasks continually, with no need for any additional training or finetuning, while obtaining strong performance both on the new task and on the old ones, a desirable ability for a robot learning from an ever increasing dataset of human experience.
To measure this ability, and highlight the difference behaviour with respect to a single language-conditioned policy, we first collect 10 demos for 3 tasks. We train the baselines on these tasks (after extracting captions via Gemini as described for the experiments of Table \ref{tab:main_table}) and evaluate them and R+X. Then, we add 10 demos of 3 new tasks, finetune the baselines, then measure performance on the 3 new tasks, and on the 3 old tasks for all methods. In Fig. \ref{fig:continual_retrieval}, we see how the performance of R3M-DiffLang and Octo deteriorates on the old task, due to the well known effect of catastrophic forgetting \cite{wang2024comprehensive}. On the other hand, R+X performance does not deteriorate, due to the ability to retrieve from an ever-growing video of tasks and adapting the behaviour model at test time on the retrieved data.

If instead we train the baselines on all the data each time, the performance does not deteriorate: however, this leads to a substantial growth in time needed to train. As the dataset grows, those networks would need an increasing amount of time per each new added task in order to be retrained, while R+X does not need any training or finetuning.

\section{Conclusion}
\label{sec:conclusion}

	We presented R+X, a method to learn robot skills from long, unlabelled videos of humans interacting with their environments. Given a language command and a live observation of the scene, R+X retrieves a sequence of relevant video clips from a long unlabelled video which it uses as context for an in-context imitation learning method that predicts what actions to execute. Through several real-world experiments on 12 everyday tasks, we demonstrate that R+X can successfully generalise to previously unseen objects, scenes, and object placements, is robust to distractor objects, and its use of VLMs for video retrieval enables R+X to generalize to complex language commands demonstrating its clear benefits over training monolithic language-conditioned policy networks. For a discussion on limitations and future work please see the appendix.

\newpage

\bibliographystyle{IEEEtran}
\bibliography{paper}

\section{Appendix}

\subsection{Additional Related Work}
In this section, we discuss some additional related papers from the literature.

\textbf{Retrieval in Robotics.} Other papers have proposed the use of retrieval in robotics. \cite{zhu2024retrievalaugmented} retrieves directly from a bank of \textit{policies} trained on robotics data. \cite{du2023behavior} retrieves from an unlabelled data of robotics experience, not human videos, and needs an additional example at test time to guide the retrieval phase, while we leverage a language command alone. \cite{dinobot} retrieves a goal observation to then perform visual servoing and replay a pre-recorded demo. R+X leverages large Vision Language Models to retrieve directly from long videos given a language query, without requiring pre-trained policies or new demonstrations as queries.

\textbf{Retrieval Augmented Generation in Language Modelling:} Large Language Models, through extensive pre-training and finetuning on web-scale datasets, are able to implicitly record countless facts and concepts in their weights \cite{openai2023gpt4, gpt4o, geminiteam2023gemini}. However, it has been proved useful to add to their implicit knowledge the ability to explicitly access datasets of information, either offline or through web search \cite{borgeaud2022improving}. This allows models to be more factually correct. In this work, we do not deal with factuality or information retrieval, but we take inspiration from these ideas to form a sort of "retrieval augmented execution", which is the main idea behind R+X. 

\textcolor{black}{
\textbf{Further Learning from Observation Papers} Recently \cite{bharadhwaj2024track2actpredictingpointtracks, yuan2024generalflowfoundationaffordance} propose methods to learn skills from videos, similarly to our paper, but focusing on optical flow between frames. \cite{bharadhwaj2024track2actpredictingpointtracks} however learns a visual-goal-conditioned policy, while we focus on language-conditioned skills extracted from a long, unlabelled video. Additionally, by focusing on optical flow of relevant objects, these methods might struggle on objects that barely move for some tasks, like pressing, as we show we can do with R+X by tracking the human hand directly. \cite{bahl2023affordanceshumanvideosversatile} learns, from in-the-wild human videos, where objects can generally be grasped and in what directions they can be moved, allowing a robot to passively learn to push, pick up, pull, etc. However, their method is more limited in the amount of skills it can learn, focusing on mostly linear movement of objects, while we show we can learn more articulated and dextereous tasks. \cite{chen2021learninggeneralizableroboticreward, Shao2020Concept2RobotLM} show that human videos can be used to learn to guide robot exploration by extracting a reward function from videos. In our method, we show we can effectively learn skill without any need for robot teleoperation or autonomous exploration. }

\subsection{Processing the Long, Unlabelled Human Video}\label{sec:hand-removal}
After recording the video of a human performing everyday tasks, we process it to remove unnecessary frames, i.e., frames that do not contain human hands. We achieve this automatically by leveraging HaMeR \cite{HaMeR} to detect the frames where hands are present. Consequently, we are left with one long, unlabelled video of smaller video clips, concatenated together, each containing only frames where a human interacts with various objects with their hands. This video is then passed to Gemini to proceed with the retrieval phase of R+X, as discussed in section~\ref{sec:retrieval}. Fig.~\ref{fig:uncut_frames} shows a short segment of the uncut video we recorded. The frames marked with red correspond to frames where no human hands were detected. The frames where human hands were detected are marked with green. After processing the video with HaMeR \cite{HaMeR} we remove the frames marked with red and are left with one long video of the green frames concatenated together.

\subsection{Stabilisation of First Person Videos}

\begin{figure*}[t!]
    \centering
            \makebox[\textwidth][c]{

    \includegraphics[width=\textwidth]{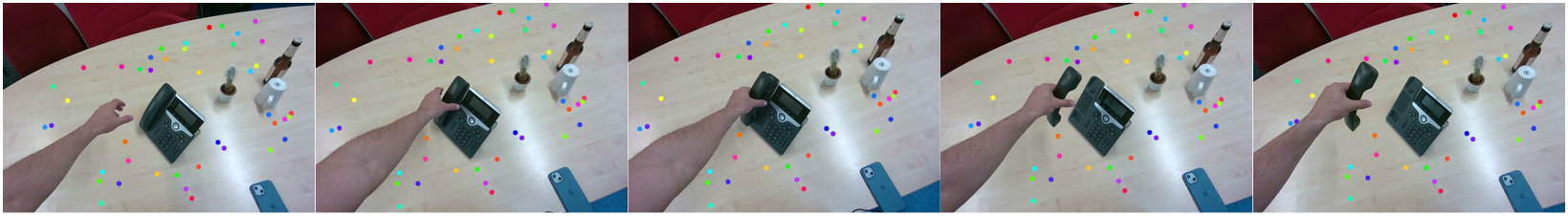}}
   
    \centering
    \caption{An example of keypoints being tracked in a retrieved video clip. Notice how, thanks to the segmentation model \cite{lüddecke2022clipseg}, we can sample keypoints that stick to static parts of the scene, like the table.}
    \label{fig:keypoints}
\end{figure*}

As we record videos using a wearable camera, the point of view of the camera changes substantially frame by frame. However, our method is based on representing the visual inputs as fixed, 3D keypoints, and the actions as a series of 3D hand joints actions expressed in the same frame of the keypoints. Therefore, we need to compute the relative movement of the camera at each step in order to express everything in the original frame of reference, the first of the recorded video clip. 

The computer vision literature has proposed several methods to address this problem \cite{matsuki2024gaussian, orbslam}. However, most of these techniques assume that the scene in front of the camera is fixed, and only the camera is moving. This is in sharp contrast with our setting, where the arms and hands of the user move in front of the camera, together with some of the objects. Classic techniques, in our experiments, failed to achieve a robust estimation of the movement of the camera, a task we hereby refer has the \textit{stabilisation of the camera}. In the following section, we describe how we stabilise each video clip we retrieve from $\mathcal{H}$ while using R+X.

In order to tackle the aforementioned problem, we leverage the use of a series of recent computer vision models. The first one, CLIPSeg \cite{lüddecke2022clipseg}, is an open-vocabulary image segmentation model. The model receives an image and a series of language queries as an input, and outputs a segmentation mask localising the pixels of the desired objects, $\text{CLIPSeg}(\mathcal{O}_{live}, [\mathcal{L}_{obj,0}, \dots, \mathcal{L}_{obj,n}]) \rightarrow M_{live} \in \mathbb{R}^{H \times W \times 1}$, where $H$ and $W$ are the height and width of the observation. We query CLIPSeg with some static parts of any scene, namely \textit{"floor", "wall", "table"}. In addition, we query it with \textit{"arm", "hand", "person"} in order to obtain segmentation masks of parts of the user arms, and remove it from the segmentation mask of the static objects. 

Given the segmentation of such static objects, we sample a set of random 2D pixel keypoints in that area. We then track the movement of such keypoints frame by frame using TAPIR \cite{doersch2023tapir}, a recent keypoint-tracking network. The position of these keypoints in 2D at each frame allows us to then project them in 3D using the depth-channel from the RGB-D camera we use. Given these sparse point-clouds, we can compute the relative $SE(3)$ rigid transformation between each new frame and the first one. We perform this process separately for each video clip where human hands where detected as discussed in section~\ref{sec:hand-removal}.

This pipeline, in our experiments, has proven to be better than classic techniques \cite{orbslam}, as it can handle the presence of the moving hands and arms of a person, together with the movement of some objects the user interacts with, as it learns to only focus on static parts of the environment, ignore the user, and track the selected keypoints over time.

\subsection{Extracting Gripper Actions from Human Hands}
\label{sec:hand-actions}
To map the human hand joints to gripper actions we deploy different heuristics based on the type of interaction the human performs in the retrieved video clips. In scenarios where a robot is equipped with an end-effector with human-like fingers, such as a dexterous 5-fingered hand, such heuristics are not needed as the hand joints would map directly to the robotic, human-like fingers.

\begin{figure*}[t!]
    \centering
        \makebox[\textwidth][c]{

    \includegraphics[width=.8\textwidth]{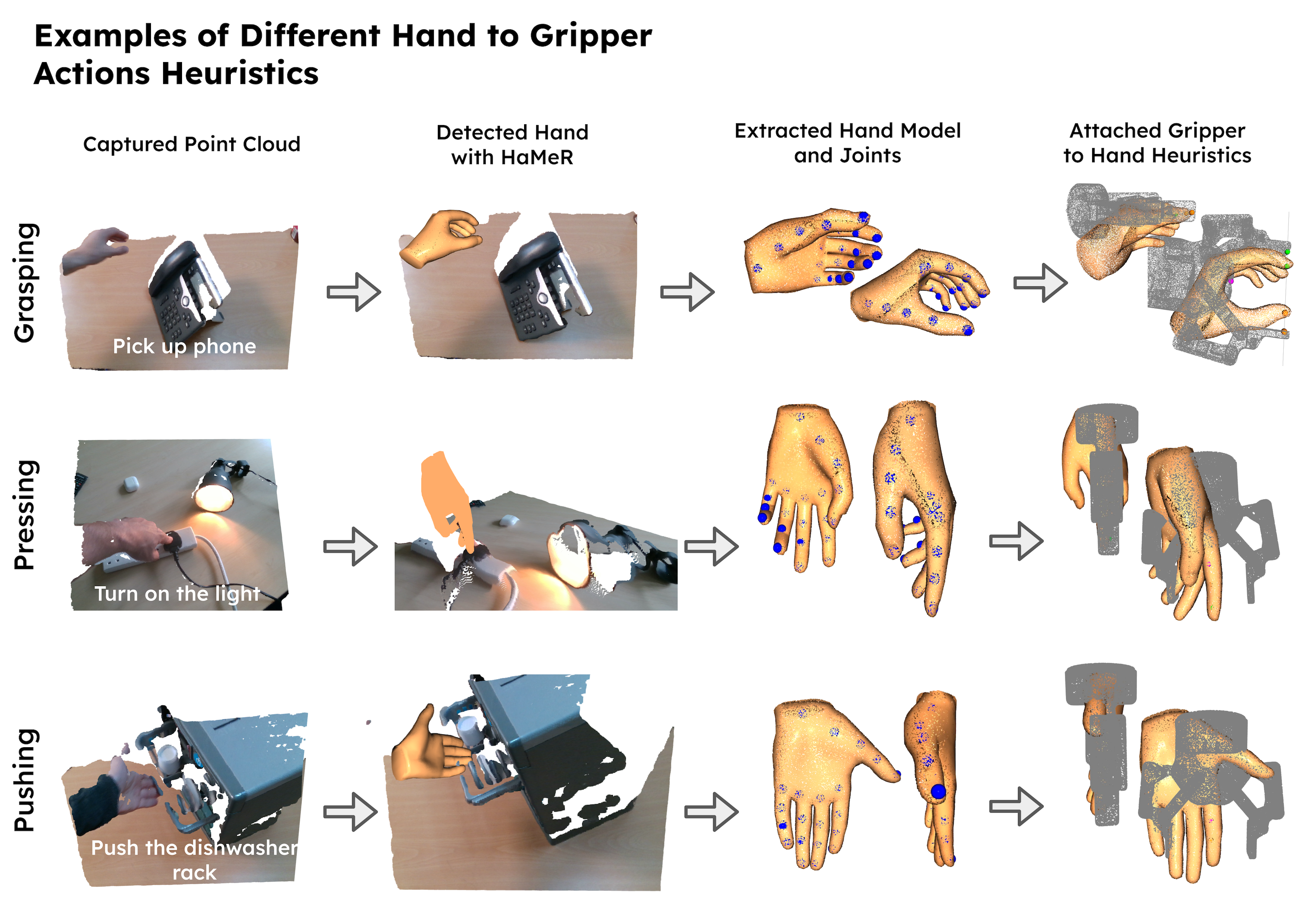}}
   
    \centering
    \caption{Examples of different human hand interactions and extracted gripper actions based on three different heuristics for grasping, pressing, and pushing. The blue dots in the third column correspond to the 21 hand joints extracted using HaMeR \cite{HaMeR}. Which heuristic to use is determined by Gemini at the Retrieval phase based on Gemini's understanding of the language command provided by the human. Note that for the pressing and pushing task the gripper is rendered as open, but in practice the gripper is closed as if it grasping something.}
    \label{fig:heuristics}
\end{figure*}

Fig.~\ref{fig:heuristics} shows instances of extracted hand joints using HaMeR \cite{HaMeR} mapped to different gripper actions for grasping, pressing, and pushing tasks. For ease of illustration, this figure demonstrates how human hand joints are mapped to gripper poses from images captured in the human video. In practice, we deploy the following heuristics on the predictions made by each method.  Our heuristics are applied to the Robotiq Gripper 2F85, but can be adapted accordingly to any other parallel jaw gripper. 

\textbf{Grasping. } (Fig.~\ref{fig:heuristics}, top) For tasks that require grasping we align spatially the tips of the gripper's fingers with the index and thumb tips of the hand. The vector joining the two tips defines an axis of rotation around which we rotate the gripper such that the bottom of the gripper is as close as possible to the middle point between the index mcp and the thumb dip. Aligning the gripper to these points enables us to obtain the gripper's pose. Consequently, for grasping tasks each method predicts, the index tip, thumb tip, index mcp and thumb dip joints. To compute the gripper close/open action, we employ an heuristic that measures the distance of the index and thumb tips and compares it to the robot gripper width. 

\textbf{Pressing. } (Fig.~\ref{fig:heuristics}, middle) For pressing tasks, we assume that the gripper is \textit{closed}. To obtain the pose of the gripper we use three points at the point of contact of the two gripper fingers when the gripper is closed. Specifically, one point is at the tip of the fingers, the other in the middle, and the last at the bottom of the gripper's fingers. We align these three points to the index tip, index pip, index mcp, and index dip of the human hand respectively. Consequently, each method predicts the index tip, index pip, index mcp, and index dip.

\textbf{Pushing. }(Fig.~\ref{fig:heuristics}, bottom) Pushing tasks, are similar to pressing tasks, where we use the same three points on the gripper but to obtain the gripper's pose, we instead align these points with the joints on the hand halfway between the index and middle finger tip, pip, mcp and dip.

For all heuristics, we match the pose of the gripper with the joints on the hand using singular value decomposition. 

Detecting which heuristic to use is done automatically by Gemini based on the human's language command during task execution. As Gemini exhibits strong semantic language understanding, it can determine whether the requested task from the human involves grasping, pressing, or pushing.

\begin{figure*}[t!]
    \centering
        \makebox[\textwidth][c]{

    \includegraphics[width=0.45\textwidth]{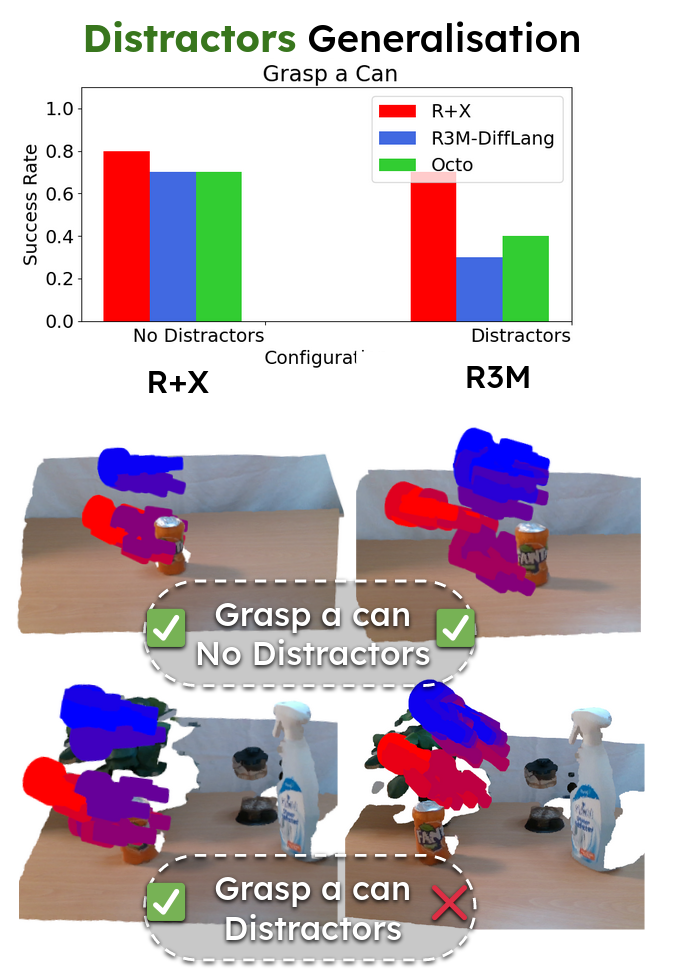}
    \includegraphics[width=0.45\textwidth]{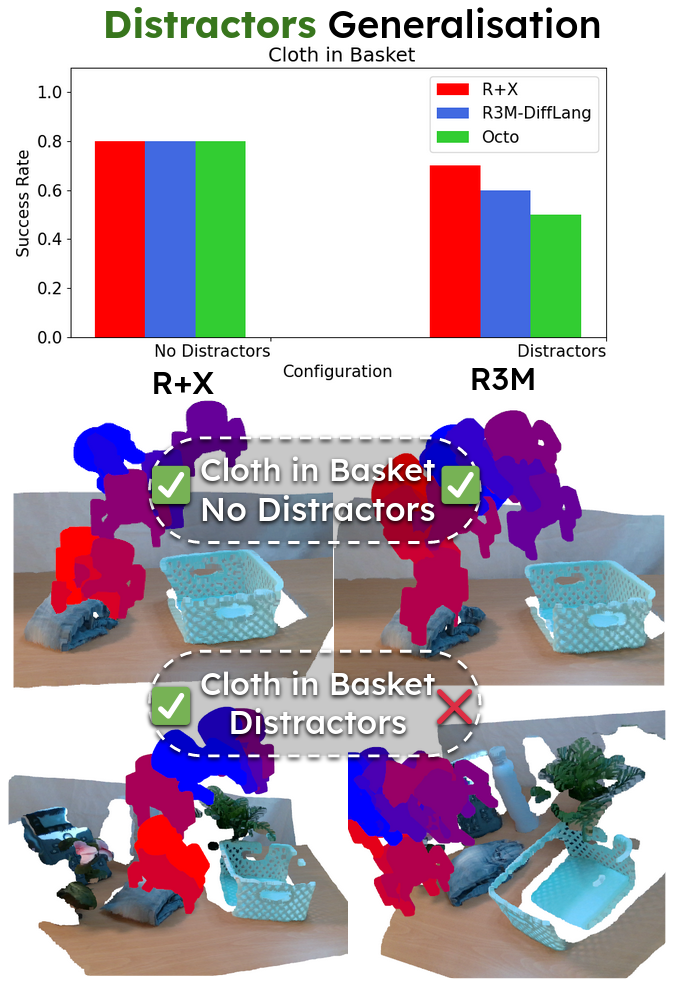}}
   
    \centering
    \caption{Examples of generalisation to distractors. Gripper moves from red to blue.}
    \label{fig:dist_gener}
\end{figure*}
\subsection{Tasks Details and Success Criteria}
We here list more details about the tasks we use in R+X, the generalisation abilities we study, and the success criteria.
\begin{itemize}
    \item \textbf{Plate in Rack:} We randomise the position of the rack on the table in a 40cm by 30cm (width, height) area and its orientation in a $[-30,30]$ degrees range. The robot starts with a plate grasped in its gripper. We place random distractors in half the runs, while half are distractors free.
    The task is successful if the plate can stand in one of the slots of the rack without falling over.
    \item \textbf{Push Rack in Dishwasher:} Due to the kinematics of the robot, we always put the dishwasher on the right side of the table, and randomise its position in a 35cm by 20cm area, and its orientation in a $[-20,20]$ degrees range. As the object is considerably large, we cannot randomise excessively its starting point or it will end out of view/out of the robot's working area. We place random distractors in half the runs, while half are distractors free.
    The task is successful if the rack is no more than 5cm outside of the dishwasher, and starts being around 15cm outside of it.
    \item \textbf{Wipe Marker from Table:} We draw a random spot with a marker on an horizontally placed whiteboard, that emulates a mark on a table, some dirt, or something spilled. We randomise the shape of the spot, that is generally $15cm^2$ in area, and its position in a 40cm by 30cm area. The robot starts with a cleaning tool grasped. We place random distractors in half the runs, while half are distractors free. The task is successful if more than 80$\%$ of the spot it erased and cleaned.
    \item \textbf{Grasp Beer}: We place a beer in a random spot on the table in a 40cm by 30cm area. In this case, we also place it 2 out of 10 times on top of other objects to test for hard spatial generalisation. This setting is better studied in the "\textit{Hard Spatial Generalisation}" subsection of the main paper. We place random distractors in half the runs, while half are distractors free. The task is solved if the robot has grasped and lifted the beer.
    \item \textbf{Cloth in Washing Machine}: We place the washing machine on the right side of the table, randomising its position in a 20cm by 20cm area, and its orientation in a $[-20,20]$ degrees range. We then place a random piece of clothing (generally clothing for kids/infants) on the left side of the table, and randomise its position in a 25cm by 25cm area, therefore testing for both spatial and object generalisation. We place random distractors in half the runs, while half are distractors free. The task is completed if the robot can place the cloth inside the hole of the washing machine, also if part is outside: as our toy washing machine is very small and its receptacle is shallow (around 15cm), it is hard also for a human to fit a cloth entirely in there. 
    \item \textbf{Open Box:} We place one of two possible boxes of different sizes in a 30cm by 30cm area on the table, and randomise its orientation in a $[-20,20]$ degrees range. We place random distractors in half the runs, while half are distractors free. The task is successful if, during the episode, the robot has partially opened the box. Opening it completely is often challenging due to kinematics constraints.
    \item \textbf{Kettle on Stove:} We place a toy stove on the table, randomising its position in a 30cm by 20cm area, and its orientation in a $[-20,20]$ degrees range. We then randomly place a random kettle in a 20cm by 20cm area and randomise its orientation in a $[-30,30]$ degrees range. By placing seen and unseen kettles, we test both for spatial and object generalisation. We place random distractors in half the runs, while half are distractors free. The task is successful if the kettle is on top of the stove, which we consider true if the intersection of the stove area and the kettle base is more than 50$\%$ than the stove area.
    \item \textbf{Close Microwave:} We place a toy microwave on the right side of the table, randomising its position in a 20cm by 20cm area and its orientation in a $[-20, 20]$ degrees range. Additionally, we open its door in a range between $[10, 45]$ degrees, We place random distractors in half the runs, while half are distractors free. We consider the task successful if the microwave door is completely closed (it has a magnetic system that keeps it closed once it is completely pushed in position).
    \item \textbf{Cloth in Basket:} We place a basket on the right side of the table in a 20cm by 20cm area, and randomise its orientation in a $[-20,20]$ degrees range. We place a random cloth (same as for the washing machine task) on the left side of the table, randomising its position in a 25cm by 25cm area.We place random distractors in half the runs, while half are distractors free. We consider the task successful if the cloth is contained in the basket, i.e. if lifting the basket, the cloth is lifted as well without falling on the table.
    \item \textbf{Pick Up Phone:} We randomly place one out of two possible phones on the table, randomising its position in a 30cm by 30cm area, and its orientation in a $[-30,30]$ degrees range. By using different phones, one of which unseen at training, we also test for object generalisation. We place random distractors in half the runs, while half are distractors free. We consider the task successful if the robot has grasped and lifted the phone from its base.
    \item \textbf{Grasp Can:} Same setting as for Grasp Beer. In this case, we also use other unseen cans at test time to test for object generalisation.
    \item \textbf{Turn On Light:} We place a light bulb and a socket on the table, randomising their positions independently in a 30cm by 30cm area, and the socket orientation in a $[-20,20]$ degrees range. We place random distractors in half the runs, while half are distractors free. The task is successful if the robot can press the light bulb's plug entirely in the socket, turning on the light. 
\end{itemize}

\begin{figure}[t!]
    \centering

    \includegraphics[width=0.5\textwidth]{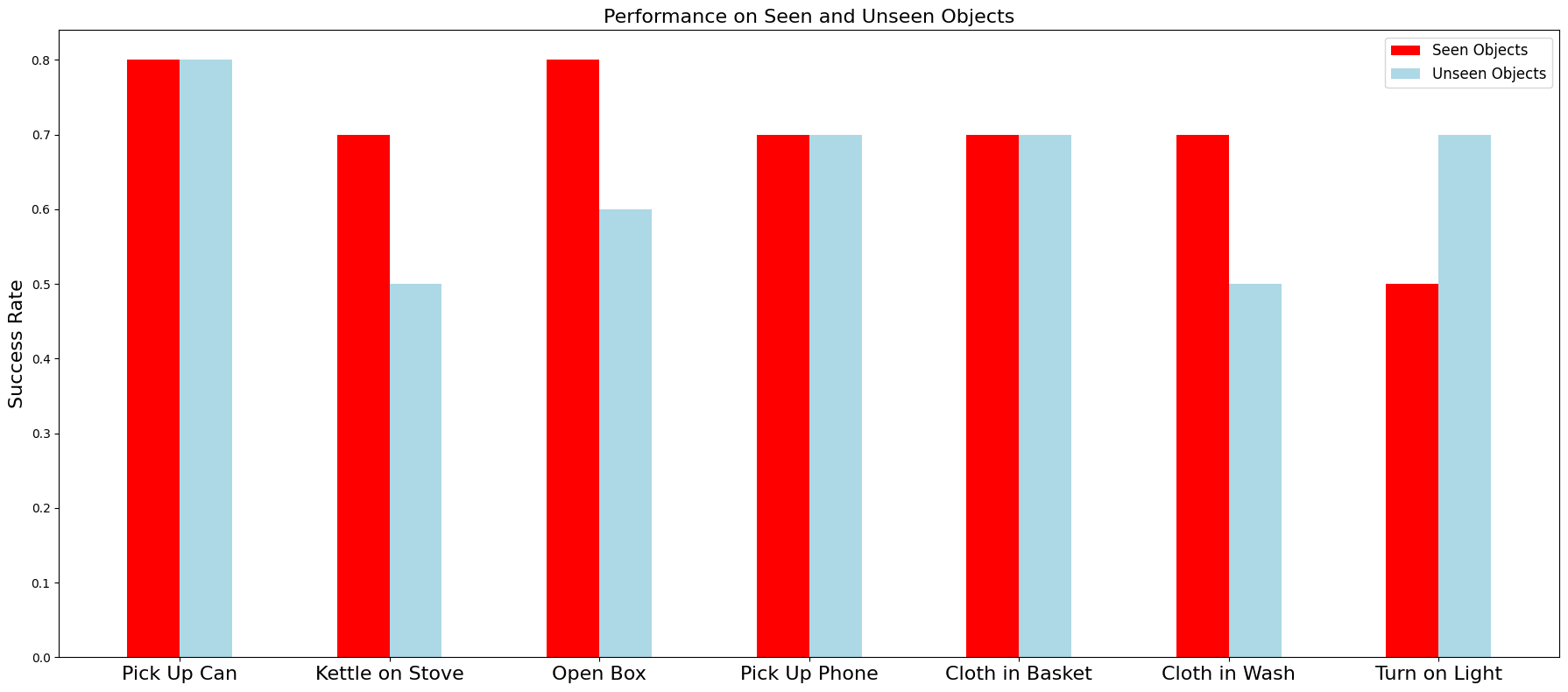}
   
    \centering
    \caption{\textcolor{black}{Success rate of R+X on seen and unseen objects from different tasks.}}
    \label{fig:unseen_obj}
\end{figure}

\begin{figure*}[t!]
    \centering

    \includegraphics[width=0.25\textwidth]{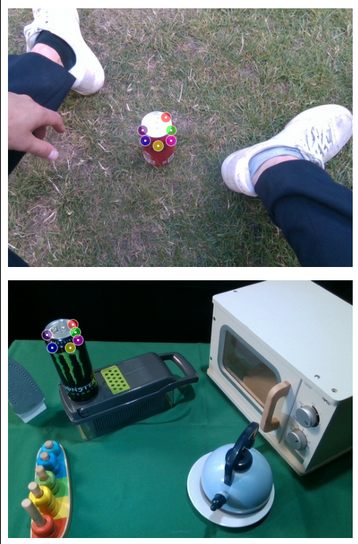}
       \includegraphics[width=0.257\textwidth]{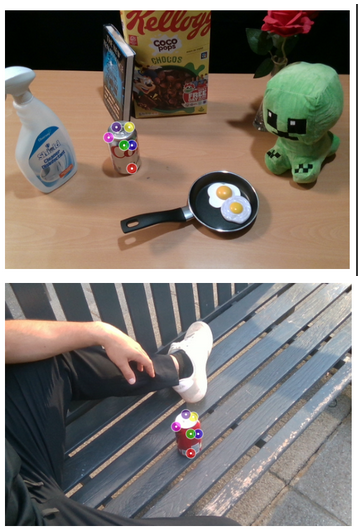}
       \includegraphics[width=0.25\textwidth]{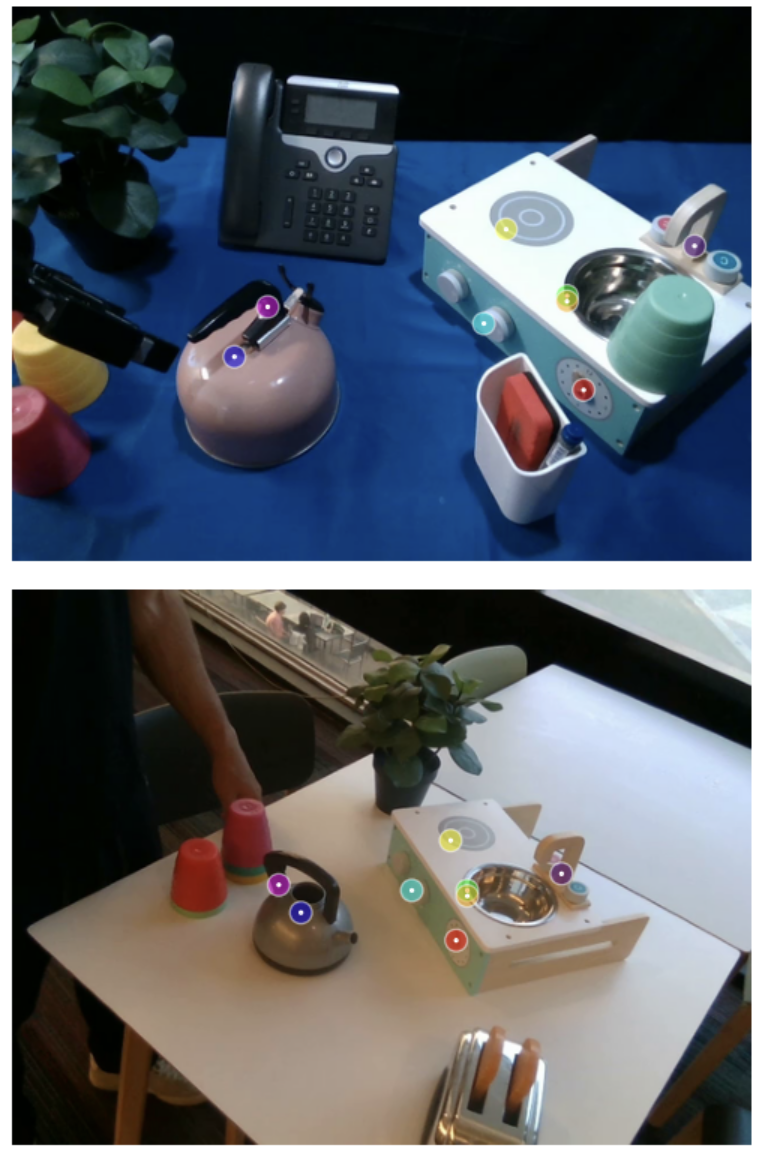}

    \centering
    \caption{\textcolor{black}{Examples of keypoints extracted for the same tasks, but with different views, settings, and target objects. Images are grouped in pairs column-wise. Keypoints are extracted between top and bottom images of each column.}}
    \label{fig:kpts-strong}
\end{figure*}

\begin{figure}[t!]
    \centering

    \includegraphics[width=0.45\textwidth]{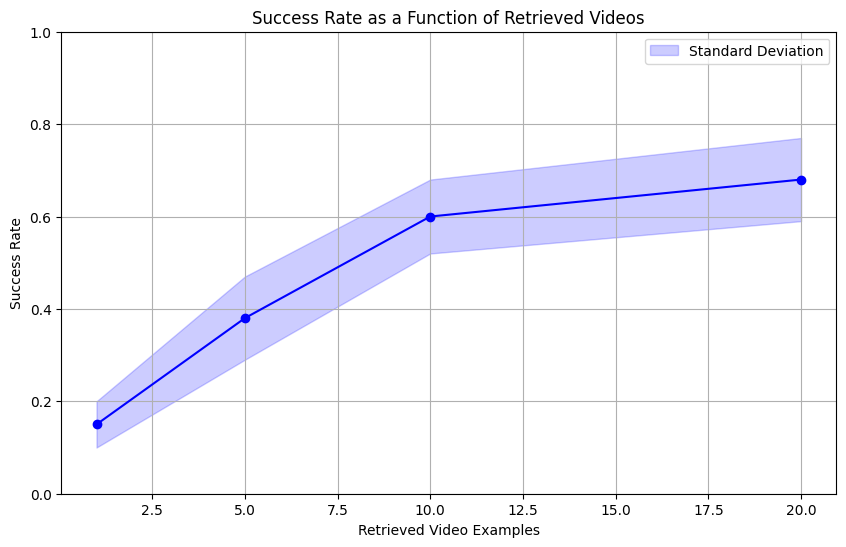}

    \centering
    \caption{\textcolor{black}{Success rate as a function of the amount of retrieved videos. We plot mean and standard deviation on 4 tasks, described below.}}
    \label{fig:in-context}
\end{figure}

\textcolor{black}{\textbf{Further Investigation: What is the performance of R+X on unseen objects?} In many of our tasks, as described before, we test R+X on unseen objects of the same categories, like unseen cans, kettles, clothes, etc. In this section, we perform a more detailed study on the performance of our method on seen and unseen objects. For the tasks that can be tested on unseen objects, we run 10 trajectories on objects seen in the human video, and 10 with unseen objects. Results in Figure \ref{fig:unseen_obj} demonstrate that our method is robust to novel, unseen objects of the same category. This is thanks to the keypoint extraction pipeline, that can extract from images a list of semantically and geometrically meaningful keypoints, that transfer across objects. Examples of such keypoints can be observed in Figure \ref{fig:kpts-strong}}.

\textcolor{black}{\textbf{How does the performance change with the number of retrieved videos?} R+X is conditioned on a set of video clips depicting the requested tasks. How many videos clips need to be retrieved from the full human video to obtain strong performance? In this experiments, we manually limit the amount of videos that are retrieved for the "\textit{pick up can, pick up telephone, cloth in basket and open box}" tasks, changing it between 1, 5, 10 and 20. We then run 10 test episodes for each of these values and compute the overall success rate. In Figure \ref{fig:in-context} we see how R+X can reach a strong performance also with as few as 10 retrieved videos.}



\textbf{Further Investigation: What are the main sources of difference in performance between R+X and a monolithic policy?}
Here we further investigate sources of differences in performance between R+X and the baselines, extending the experimental investigation of the main paper. In particular, here we focus on the presence of distractors.

\textbf{\textit{Distractors Generalisation}}: One of the main differences between R+X and the baselines is that the former, after retrieving a set of video clips from $\mathcal{H}$ depicting the requested task, can extract a set of keypoints that are semantically and geometrically meaningful for the objects to interact with, as described in the main paper. These keypoints are generally robust to distractors, as they focus only on the DINO features that are common in all images, whereas distractors generally through out the human video $\mathcal{H}$ and the execution. The baselines, being monolithic policies, receive as input the observation as it is, and must learn to generalise to the presence of distractors during the long training phase.

To evaluate the effect of distractors on the scene, we test R+X and the baselines on two tasks, "\textit{grasp a can}" and "\textit{put the cloth in the basket}", emulating the experimental scenario of the "\textit{Hard Generalisations}" sections. We run 10 runs without distractors, and 10 with, explicitly measuring the performance in the two cases. We can see in the results of Fig. \ref{fig:dist_gener} how R+X is more robust to the presence of random distractor objects, highlighting the advantage of extracting semantically and geometrically meaningful keypoints \textit{after} having received a language command and having retrieved the corresponding video clips.

\subsection{Further Details on Baselines:}
\subsubsection{Octo}
One of baselines that we have compared against is a fine tuned version of Octo \cite{octomodelteam2024octo}. We have used the original code provided by the authors and adapted it for the specific input and output relevant to this work. More specifically, we designed the input of the model to be the RGB image of size $256 \times 256$ of the live observation, $\mathcal{O}_{live}$ and a language description of the task $\mathcal{L}$. The output, instead, is $\mathcal{J}_{live}$, the trajectory of desired hand joints 3D points to be translated to gripper poses as described before. Unlike KAT, these methods require a fixed length of the output trajectory. We therefore pad it to 40 times steps. The output is ultimately a tensor of dimension 480, which corresponds to a flattened tensor of dimension $(40, 4, 3)$. The 4 predicted hand joints are described in section \ref{sec:hand-actions} along with how they are mapped to robot actions.

Having set the input and output dimensions, as well as modalities, the model has been fine tuned starting from the provided checkpoint "octo-small-1.5" and it was trained until convergence.

\subsubsection{R3M-DiffLang}
The other baseline we have compared against has been the combination of R3M \cite{nair2022r3m} and \cite{chi2023diffusion}, with the inclusion of language conditioning. The resulting model leveraged the ResNet-50 variant of R3M to encode the $256 \times 256$ live observation RGB image $\mathcal{O}_{live}$ and Sentence-BERT \cite{reimers-2019-sentence-bert} to encode the language description of the task $\mathcal{L}$. The resulting feature vectors for the image and language description was then concatenated and passed to a UNet \cite{ronneberger2015unet} trained via diffusion following the original code of \cite{chi2023diffusion}. The output of the diffusion head has the same size as the output of Octo, as we pad the trajectory $\mathcal{J}_{live}$ to 40 timesteps. In order to take advantage of the pre-trained vision knowledge of R3M, the encoder has been initialised with the provided weights and kept frozen, as it has been done for the language encoder. The diffusion policy head has been trained from scratch using our own dataset of trajectories until convergence was reached. 

\subsection{Additional Discussion on Limitations:}
 While R+X can learn a large range of everyday tasks with high success rate directly from everyday human videos and no additional training, it still has some limitations we discuss here and aim to address as part of future work.  Firstly, the errors arising from the hand pose prediction currently inhibits us from learning precise tasks like methods that spend a long time collecting, on-board task-specific data \cite{papagiannis2024miles, luo2024serl}. Furthermore, while the keypoints extraction leads to strong spatial generalisation properties, it can become a bottleneck in the presence of many similar objects. In future works, methods that find a dynamic number of keypoints could tackle this issue. Finally, due to the use of Foundation Models for retrieval and execution, our method has a few seconds delay between the issuing of the command and execution, that however has reduced dramatically over the last months with the release of faster models such as GPT-4o \cite{gpt4o} or Gemini Flash \cite{gemini15}.

\begin{figure*}[b]
    \centering
        \vspace{-5ex}\makebox[0.8\textwidth][c]{

    \includegraphics[width=\textwidth]{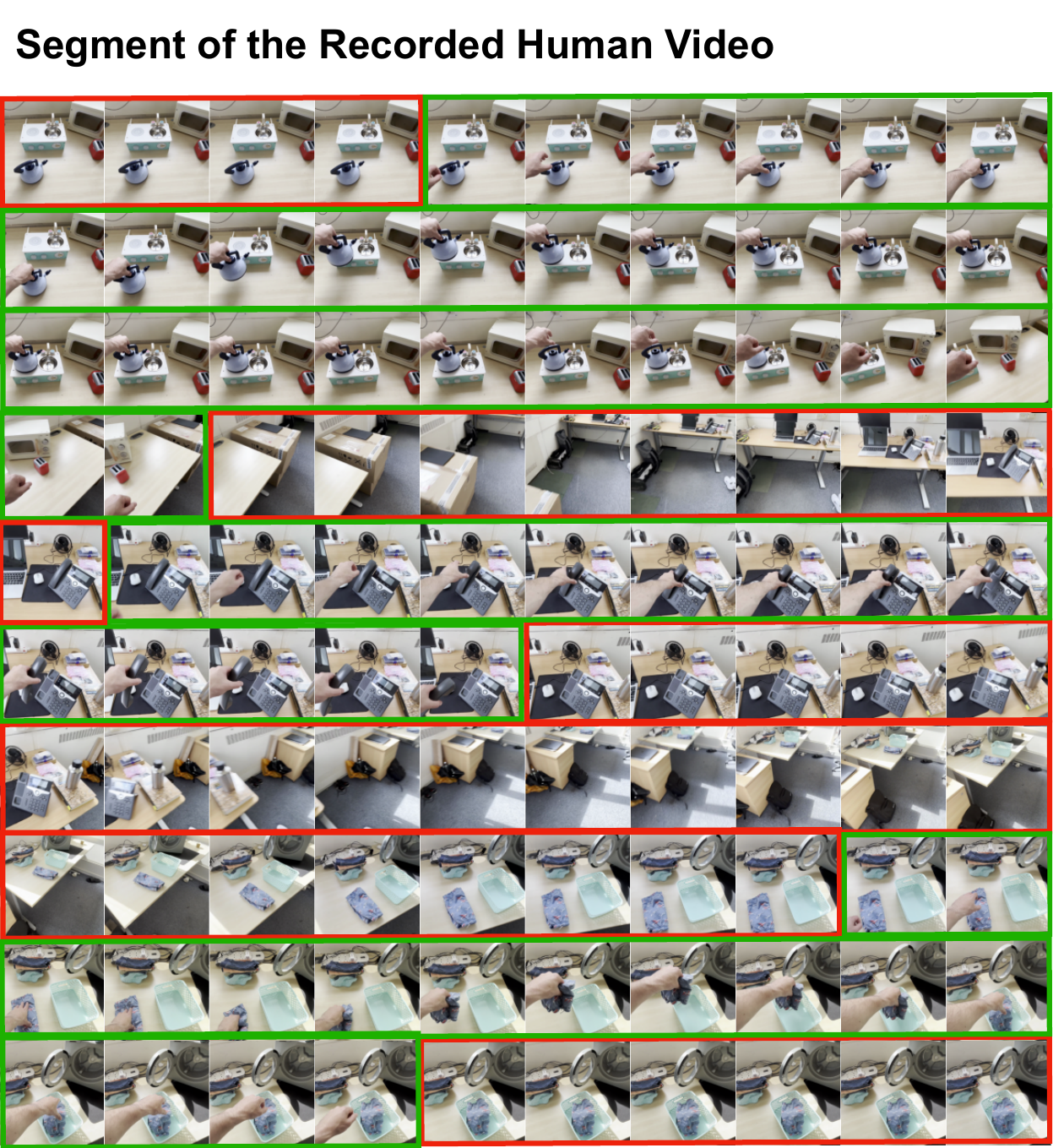}}
   
    \centering
    \caption{A short segment of our recorded long, unlabelled video of a human performing everyday tasks. The frames marked with red correspond to frames where no human hands were detected, while the frames marked with green correspond to frames where HaMeR \cite{HaMeR} detected human hands. In practice, we discard the frames where no human hands are present and only retain a long video of frames concatenated together depicting only the human interacting with various objects.}
    \label{fig:uncut_frames}
\end{figure*}


\end{document}